Building Machines that Learn and Think for Themselves:

Commentary on Lake, Ullman, Tenenbaum, and Gershman,

*Behavioral and Brain Sciences,* 2017

M. Botvinick, D.G.T. Barrett, P. Battaglia, N. de Freitas, D. Kumaran, J. Z Leibo, T. Lillicrap, J. Modayil, S. Mohamed, N.C. Rabinowitz, D. J. Rezende, A. Santoro, T. Schaul, C. Summerfield, G. Wayne, T. Weber, D. Wierstra, S. Legg, and D. Hassabis

DeepMind

DeepMind
6 Pancras Square
Kings Cross
London, N1c4AG

Botvinick: botvinick@google.com, deepmind.com

Abstract:

We agree with Lake and colleagues on their list of 'key ingredients' for building humanlike intelligence, including the idea that model-based reasoning is essential. However, we favor an approach that centers on one additional ingredient: autonomy. In particular, we aim toward agents that can both build and exploit their own internal models, with minimal human hand-engineering. We believe an approach centered on autonomous learning has the greatest chance of success as we scale toward real-world complexity, tackling domains for which ready-made formal models are not available. Here we survey several important examples of the progress that has been made toward building autonomous agents with humanlike abilities, and highlight some outstanding challenges.

Lake and colleagues identify some extremely important desiderata for humanlike intelligence. We agree with many of their central assertions: Humanlike learning and decision making surely do depend upon rich internal models; the learning process must be informed and constrained by prior knowledge, whether this is part of the agent's initial endowment or acquired through learning; and naturally, prior knowledge will offer the greatest leverage when it reflects the most pervasive or ubiquitous structures in the environment, including physical laws, the mental states of others, and more abstract regularities such as compositionality and causality. Together, these points comprise a powerful set of target goals for AI research. However, while we concur on these goals, we choose a differently calibrated strategy for accomplishing them. In particular, we favor an approach that prioritizes autonomy, empowering artificial agents to learn their own internal models and how to use them, mitigating their reliance on detailed configuration by a human engineer.

Lake and colleagues characterize their position as "agnostic with regards to the origins of the key ingredients" of humanlike intelligence. This agnosticism implicitly licenses a modeling approach in which detailed, domain-specific information can be imparted to an agent directly, an approach for which some of the authors' Bayesian Program Learning (BPL) work is emblematic. The two domains Lake and colleagues focus most upon – physics and theory of mind – are amenable to such an approach, in that these happen to be fields for which mature scientific disciplines exist. This provides unusually rich support for hand design of cognitive models. However, it is not clear that such hand design will be feasible in other more idiosyncratic domains where comparable scaffolding is unavailable. Lake and colleagues (2015) were able to extend the approach to Omniglot characters by intuiting a suitable (stroke-based) model, but are we in a position to build comparably detailed domain models for such things as human dialogue or architecture? What about Japanese cuisine or ice skating? Even video-game play appears daunting, when one takes into account the vast amount of semantic knowledge that is plausibly relevant (knowledge about igloos, ice floes, cold water, polar bears, video-game levels, avatars, lives, points, and so forth). In short, it is not clear that detailed knowledge engineering will be realistically attainable in all areas we will want our agents to tackle.

Given this observation, it would appear most promising to focus our efforts on developing learning systems that can be flexibly applied across a wide range of domains, without an unattainable overhead in terms of a priori knowledge. Encouraging this view, the recent machine learning literature offers many examples of learning systems conquering tasks that had long eluded more hand-crafted approaches, including object recognition, speech recognition, speech generation, language translation, and (significantly) game play (Silver et al., 2016b). In many cases, such successes have depended on large amounts of training data, and have implemented an essentially model-free approach. However, a growing volume of work suggests that flexible, domain-general learning can also be successful on tasks where training data are scarcer and where model-based inference is important.

For example, Rezende and colleagues (2016) reported a deep generative model that produces plausible novel instances of Omniglot characters after one presentation of a model character, going a significant distance toward answering Lake's "Character Challenge." Lake and colleagues call attention to this model's "need for extensive pretraining." However, it is not clear why their pre-installed model is to be preferred over knowledge acquired through pretraining. In weighing this point, it is important to note that the human modeler, in order to furnish the BPL architecture with its "startup software," must draw on his or her own large volume of prior experience. In this sense, the resulting BPL model is dependent on the human designer's own 'pretraining.'

A more significant aspect of the Rezende model is that it can be applied without change to very different domains, as Rezende and colleagues (2016) demonstrate through experiments on human facial images. This flexibility is one hallmark of an autonomous learning system, and contrasts with the more purpose-built flavor of the BPL approach, which relies on irreducible primitives with domain-specific content (e.g., the strokes in Lake's Omniglot model). Furthermore, a range of recent work with deep generative models (e.g. van den Oord, 2016; Ranzato et al., 2016) indicates that they can identify quite rich structure, increasingly avoiding silly mistakes like those highlighted in Lake and colleagues' Figure 6.

Importantly, a learning-centered approach does not prevent us from endowing learning systems with some forms of *a priori* knowledge. Indeed, the current resurgence in neural network research was triggered largely by work that does just this, for example by building an assumption of translational invariance into the weight matrix of image classification networks (Krizhevsky et al., 2012). The same strategy can be taken in order to endow learning systems with assumptions about compositional and causal structure, yielding architectures that learn efficiently about the dynamics of physical systems, and even generalize to previously unseen numbers of objects (Battaglia et al., 2016), another challenge problem highlighted by Lake and colleagues. In such cases, however, the inbuilt knowledge takes a highly generic form, leaving wide scope for learning to absorb domain-specific structure (see also Eslami et al, 2016; Raposo et al., 2016; Reed and de Freitas, 2016).

Under the approach we advocate, high-level prior knowledge and learning biases can be installed not only at the level of representational structure, but also through larger-scale architectural and algorithmic factors, such as attentional filtering (Eslami et al., 2016), intrinsic motivation mechanisms (Bellemare et al, 2016), or episodic learning (Blundell et al., 2016). Recently developed architectures for memory storage (e.g., Graves et al., 2016) offer a critical example. Lake and colleagues describe neural networks as implementing "learning as a process of gradual adjustment of connection strengths." However, recent work has introduced a number of architectures within which learning depends on rapid storage mechanisms, independent of connection-weight changes (Duan et al., 2016; Graves et al., 2016; Wang et al., 2016; Vinyals et al., 2016). Indeed, such mechanisms have even been applied to one-shot classification of Omniglot characters (Santoro et al., 2016) and Atari video game play (Blundell et al., 2016). Furthermore, the connection-weight changes that do occur in such models can serve in part to support learning-to-learn (Duan et al., 2016; Graves et al., 2016; Ravi and Larochelle, 2017; Vinyals et al., 2016; Wang et al., 2016;), another of Lake and colleagues' key ingredients for humanlike intelligence. As recent work has shown (Duan et al., 2016; Hochreiter et al., 2001; Santoro et al., 2016; Wang et al., 2016;), this learning-to-learn mechanism can allow agents to adapt rapidly to new problems, providing a novel route to install prior knowledge through learning rather than by hand.

Another reason why we believe it may be advantageous to autonomously learn internal models is that such models can be shaped directly by specific, concrete tasks. A model is valuable not because it veridically captures some ground truth, but because it can be efficiently leveraged to support adaptive behavior. Just as Newtonian mechanics is sufficient for explaining many everyday phenomena, yet too crude to be useful to particle physicists and cosmologists, an agent's models should be calibrated to its tasks. This is essential for models to scale to real-world complexity, since it is usually too expensive, or even impossible, for a system to acquire and work with extremely fine-grained models of the world (Botvinick & Weinstein, 2015; Silver et al., 2016a). Of course, a good model of the world should be applicable across a range of task conditions, even ones that have not been previously encountered. However, this simply implies that models should be calibrated not only to individual tasks, but to the distribution of tasks -- inferred through experience or evolution -- that is likely to arise in practice.

Finally, in addition to the importance of model-building, it is important to recognize that real autonomy also depends on control functions, the processes that leverage models in order to make actual decisions. An autonomous agent needs good models, but it also needs to know how to make use of them (Botvinick & Cohen, 2014), especially in settings where task goals may vary over time. This point also favors a learning- and agent-based approach, since it allows control structures to co-evolve with internal models, maximizing their compatibility. Though efforts to capitalize on these advantages in practice are only in their infancy, recent work from Hamrick and colleagues (2017), which simultaneously trained an internal model and a corresponding set of control functions, provides a case study of how this might work.

Our comments here, like the target article, have focused on model-based cognition. However, an aside on model-free methods is warranted. Lake and colleagues describe model-free methods as providing peripheral support for model-based approaches. However, there is abundant evidence that model-free mechanisms play a pervasive role in human learning and decision making (Kahneman, 2011). Furthermore, the dramatic recent successes of model-free learning in areas such as game play, navigation, and robotics suggest that it may constitute a first-class, independently valuable approach for machine learning. Lake and colleagues call attention to the heavy data demands of model-free learning, as reflected in DQN learning curves. However, even since the initial report on DQN (Mnih et al., 2015), techniques have been developed that significantly reduce the data requirements of this and related model-free learning methods, including prioritized memory replay (Schaul et al., 2015), improved exploration methods (Bellemare et al., 2016), and techniques for episodic reinforcement learning (Blundell et al., 2016). Given the pace of such advances, it may be premature to relegate model-free methods to a merely supporting role.

To conclude, despite the differences we have focused on here, we agree strongly with Lake and colleagues that humanlike intelligence depends at least in part on richly structured internal models. Our approach to building humanlike intelligence can be summarized as a commitment to developing autonomous agents: agents that shoulder the burden of building their own models and arriving at their own procedures for leveraging them. Autonomy, in this sense, confers a capacity to build economical task-sensitive internal models and to adapt flexibly to diverse circumstances, while avoiding a dependence on detailed, domain-specific prior information. A key challenge in pursuing greater autonomy is the need to find more efficient means of extracting knowledge from potentially limited data. But recent work on memory, exploration, compositional representation, and processing architectures provides grounds for optimism. In fairness, the authors of the target article have also offered, in other work, some indication of how their approach might be elaborated to support greater agent autonomy (Lake et al., 2017). We may thus be following slowly converging paths. On a final note, it is worth pointing out that as our agents gain in autonomy, the opportunity increasingly arises for us to obtain new insights from what they themselves discover. In this way, the pursuit of agent autonomy carries the potential to transform the current AI landscape, revealing new paths toward humanlike intelligence.

References

Battaglia, P., Pascanu, R., Lai, M. and Rezende, D.J. (2016). Interaction networks for learning about objects, relations and physics. In *Advances in Neural Information Processing Systems*, 4502-4510.

Bellemare, M. G., Srinivasan, S., Ostrovski, G., Schaul, T., Saxton, D. and Munos, R. (2016). Unifying count-based exploration and intrinsic motivation." *arXiv preprint arXiv:1606.01868*.

Blundell, C., Uria, B., Pritzel, A., Li, Y., Ruderman, A., Leibo, J. Z., Rae, J., Wierstra, D. and Demis Hassabis (2016). Model-free episodic control. *arXiv preprint, arXiv:1606.04460*.

Botvinick, M. M., and Cohen, J. D. The computational and neural basis of cognitive control: charted territory and new frontiers. *Cognitive science, 38*, 1249-1285.

Botvinick, M., Weinstein, A., Solway, A., & Barto, A. (2015). Reinforcement learning, efficient coding, and the statistics of natural tasks. *Current Opinion in Behavioral Sciences*, *5*, 71-77.

Duan, Y., Schulman, J., Chen, X., Bartlett, P. L., Sutskever, I., and Abbeel, P. (2016) $RL^2$: Fast reinforcement learning via slow reinforcement learning." *arXiv preprint arXiv:1611.02779*.

Eslami, S. M., Heess, N., Weber, T., Tassa, Y., Kavukcuoglu, K., and Hinton, G. E. "Attend, Infer, Repeat: Fast Scene Understanding with Generative Models." *arXiv preprint arXiv:1603.08575* (2016).

Graves, A., Wayne, G., Reynolds, M., Harley, T., Danihelka, I., Grabska-Barwińska, A., Colmenarejo, S.G., Grefenstette, E., Ramalho, T., Agapiou, J. and Badia, A.P. (2016). Hybrid computing using a neural network with dynamic external memory. *Nature*, *538*, 471-476.

Hamrick, J. B., Ballard, A. J., Pascanu, R., Vinyals, O., Heess, N., Battaglia, P. W. (2016). Metacontrol for adaptive imagination-based optimization. Under open review: https://openreview.net/pdf?id=Bk8BvDqex.

Hochreiter, S. A., Younger, S., and Conwell, P. R. (2001). Learning to learn using gradient descent. In *International Conference on Artificial Neural Networks*, pp. 87-94. Springer Berlin Heidelberg.

Kahneman, D. (2011). *Thinking, fast and slow*. Macmillan.

Krizhevsky, A., Sutskever, I., and Hinton, G. E.. (2012). Imagenet classification with Botvinick and colleagues                                                                                             Commentary deep convolutional neural networks. In *Advances in neural information processing systems*, 1097-1105.

Lake, B. M., Lawrence, N. D., and Tenenbaum, J. B. (2016). The emergence of organizing structure in conceptual representation." *arXiv preprint arXiv:1611.09384*.

Lake, B. M., Salakhutdinov, R., & Tenenbaum, J. B. (2015). Human-level concept learning through probabilistic program induction. *Science*, *350*, 1332-1338.

Mnih, V., Kavukcuoglu, K., Silver, D., Rusu, A.A., Veness, J., Bellemare, M.G., Graves, A., Riedmiller, M., Fidjeland, A.K., Ostrovski, G. and Petersen, S., 2015. Human-level control through deep reinforcement learning. *Nature*, *518*, 529-533.

van den Oord, A., Kalchbrenner, N. and Kavukcuoglu, K. (2016). Pixel recurrent neural networks. *arXiv preprint arXiv:1601.06759*.
Vinyals, O., Blundell, C., Lillicrap, T., and Wierstra, D. (2016). Matching networks for one shot learning. In *Advances in Neural Information Processing Systems*, pp. 3630-3638.

Raposo, D., Santoro, A., Barrett, D. G. T., Pascanu, R., Lillicrap, T., Battaglia, P. (2017) Discovering objects and their relations from entangled scene representations (Under open review: //openreview.net/pdf?id=Bk2TqVcxe)

Ravi, S. and Larochelle, H. (2017). Optimization as a model for few-shot learning. (Under open review: https://openreview.net/pdf?id=rJY0-Kcll).

Rezende, D. J., Mohamed, S., Danihelka, I., Gregor, K., & Wierstra, D. (2016). One-Shot Generalization in Deep Generative Models. *arXiv preprint arXiv:1603.05106*.

Ranzato, M., Szlam, A., Bruna, J., Mathieu, M., Collobert, R., Chopra, S. (20160 Video (language) modeling: a baseline for generative models of natural videos. *arXiv preprint arXiv:1412.6604*.

Reed, S. and de Freitas, N. (2016). Neural programmer-interpreters. *ICLR*.

Santoro, A., Bartunov, S., Botvinick, M., Wierstra, D., and Lillicrap, T. (2016). One-shot learning with memory-augmented neural networks." *arXiv preprint arXiv:1605.06065*.

Schaul, T., Quan, J., Antonoglou, I., and Silver, D. (2015). Prioritized experience replay." *arXiv preprint arXiv:1511.05952*.

Silver, D., van Hasselt, H., Hessel, M., Schaul, T., Guez, A., Harley, T., Dulac-Arnold, G. et al. (2016a) The predictron: End-to-end learning and planning. *arXiv preprint arXiv:1612.08810*.

Silver, D., Huang, A., Maddison, C.J., Guez, A., Sifre, L., Van Den Driessche, G.,

Schrittwieser, J., Antonoglou, I., Panneershelvam, V., Lanctot, M. and Dieleman, S., (2016b). Mastering the game of Go with deep neural networks and tree search. *Nature*, *529*, 484-489.

Wang, J. X., Kurth-Nelson, Z., Tirumala, D., Soyer, H., Leibo, J. Z., Munos, R., Blundell, C., Kumaran, D. and Botvinick, M. (2016). Learning to reinforcement learn. *arXiv preprint arXiv:1611.05763*.